\documentclass[a4paper]{article}
\usepackage{times}
\usepackage{INTERSPEECH2019}
\usepackage{graphicx}
\usepackage{amsfonts}
\usepackage{amssymb}
\usepackage{amsmath}
\usepackage{floatrow}
\usepackage{bm}
\usepackage{graphicx}
\usepackage{subfigure}
\usepackage{bbm}

\title{Learning Alignment for Multimodal Emotion Recognition from Speech
}

\name{ 
    Haiyang Xu$^1$, 
    Hui Zhang$^1$,
    Kun Han$^2$,
    Yun Wang$^3$,
    Yiping Peng$^1$,
    Xiangang Li$^1$
}

\address{
  $^1$DiDi Chuxing, Beijing, China\\
  $^2$DiDi Research America, Mountain View, CA, USA\\
  $^3$Peking University, Beijing, China
  }
\email{\{xuhaiyangsnow, ethanzhanghui, kunhan\}@didiglobal.com,\\
wangyunazx@pku.edu.cn,
\{pengyiping, lixiangang\}@didiglobal.com}

\begin{document}
\maketitle

\begin{abstract}
Speech emotion recognition is a challenging problem because human convey emotions in subtle and complex ways. For emotion recognition on human speech, one can either extract emotion related features from audio signals or employ speech recognition techniques to generate text from speech and then apply natural language processing to analyze the sentiment. Further, emotion recognition will be beneficial from using audio-textual multimodal information, it is not trivial to build a system to learn from multimodality. One can build models for two input sources separately and combine them in a decision level, but this method ignores the interaction between speech and text in the temporal domain. In this paper, we propose to use an attention mechanism to learn the alignment between speech frames and text words, aiming to produce more accurate multimodal feature representations. The aligned multimodal features are fed into a sequential model for emotion recognition. 
We evaluate the approach on the IEMOCAP dataset and the experimental results show the proposed approach achieves the state-of-the-art performance on the dataset. \footnote{Our code is available at https://github.com/didi/delta}


\noindent\textbf{Index Terms}: Emotion Recognition, Multimodal, Attention, Alignment
\end{abstract}

\section{Introduction}

Despite the tremendous progress made in speech and natural language understanding in recent years, we are still far from being able to naturally interact with machines. Building a system to understand human emotions is paramount for many human-computer interaction applications. However, it is very challenging to build such systems. 

Human express emotion through various modalities such as voice, facial expression, body posture, therefore utilizing multiple modalities may accurately capture expressed emotion and lead to better recognition results than unimodal approaches \cite{ngiam2011multimodal}. Many studies focused on using audio-visual modalities for emotion recognition, because both are very informative features on emotional expression. However, in many real applications, it is not feasible to access audio-visual data and only audio data is available, for example, emotion recognition for call centers or fatigue detection for drivers. In this case, an emotion recognition system only using speech signals is favorable.

In daily life, human utter a sentence in a natural way which conveys emotion states through both voice and contents. Although there are many studies on emotion recognition in speech and sentiment analysis in text, only a few study considered doing them jointly. Furthermore, in the scenarios where only speech data is accessible, one can utilize the automatic speech recognition (ASR) technique to convert audio signals into text and then apply a multimodal model to learn emotion from speech and text simultaneously. In this way, text data are created by an ASR system, which is usually trained from another large amount of dataset for a speech recognition purpose. Therefore, it is arguably that we employ prior knowledge learned from another dataset for the emotion recognition task. This can be considered as a transfer learning scheme, similar to pretraining word embedding in natural language processing (NLP) \cite{mikolov2013distributed} or pretraining models on ImageNet for object recognition \cite{russakovsky2015imagenet}.

To effectively utilize both speech and text data, one needs to design a model to jointly learn features from different domains. Although some studies combined both features and trained a multimodal model, few work focused on the temporal relationship between speech and text in a fine-grained level. We believe that, since the speech and text inherently co-exist in the temporal dimension, a multimodal system will be benefit from using the alignment information. In fact, in an end-to-end speech recognition system, the model employs an attention mechanism to have a decoded word to attend to its corresponding speech frames \cite{chorowski2015attention,chan2016listen}. Inspired by this work, we utilize an attention network to learn the alignment between speech and text. The aligned speech and text features are combined in the word level and serve as multimodal features for an emotional utterance. We then use a recurrent network, for example a long short-term memory (LSTM) network, to model the sequence for emotion recognition. We emphasize that, although an ASR system can output an alignment result (i.e., hard alignment for hidden Markov model based systems, and soft alignment for attention-based systems), our approach does not require the alignment from ASR. The alignment is completely learned from the attention mechanism in the model. There are two advantages for using the learned alignment: first, our approach is suitable for the scenario where an ASR system is a black box and can only output the recognized text, for example, using Google speech recognition API; second, the alignment is learned for an emotion recognition purpose and may be better than the alignment from speech recognition. 

In the next section, we relate our work to prior emotion recognition studies. We then describe our proposed approach in detail in Section 3. We show the experimental results in Section 4 and conclude the paper in Section 5.

\section{Related Work}
Machine learning technology has been used to resolve speech emotion recognition problems for decades. Previous studies usually extracted engineered low-level features or high-level statistical features and applied a classifier for emotion recognition, such as Gaussian mixture models \cite{neiberg2006emotion}, hidden Markov model \cite{nogueiras2001speech}, support vector machines \cite{mower2011framework}, neural networks \cite{stuhlsatz2011deep, kim2013emotion}. 

Recent studies on deep learning have shown that neural networks are capable of learning high-level features from raw data and increasing studies attempted to build systems using neural architectures. In \cite{han2014speech}, researchers demonstrated the effectiveness of emotional feature learning using deep neural networks (DNNs). Some studies employed recurrent neural networks (RNNs) for emotion recognition due to the sequential structure of speech signals, such as \cite{lee2015high,mirsamadi2017automatic,sarma2018emotion,li2018attention}. In addition, since convolutional neural networks (CNNs) are designed to learn local spatial features which are suitable for feature extraction in the spectral domain, some studies utilized CNNs to extract features and combined with a sequential model, for example, LSTMs \cite{trigeorgis2016adieu,satt2017efficient}. 

Multimodal learning is an important topic in machine learning \cite{ngiam2011multimodal}. In emotion recognition, many studies extracted features from audio, visual, or textual domains and then fuse them either in the feature levels or decision levels \cite{busso2004analysis,wollmer2010context,poria2017review}. To leverage information from speech signals and text sequences, previous study \cite{yoon2018multimodal} used neural networks to model two sequences separately and use direct concatenation of two modalities for emotion classification. In \cite{zadeh2017tensor}, a tensor fusion network was proposed to fuse features from different modalities and learn intra-modality and inter-modality dynamics. In \cite{poria2017context}, an LSTM-based model was utilized to learn contextual information from the utterances for sentiment analysis.

Attention networks are also related to our work. In \cite{bahdanau2014neural}, an attention network was firstly proposed to align the input and the output sequences for machine translation in NLP. Following this study, researchers in the speech area adopted the idea and utilized the attention mechanism for end-to-end speech recognition \cite{chorowski2015attention,chan2016listen}. In speech emotion recognition, several studies have been used attention networks \cite{mirsamadi2017automatic,sarma2018emotion}, however, they mainly utilized attention only for sequential modeling. To our knowledge, our work is the first work utilizing it to align speech and text sequences.


\section{Algorithm Details}
\begin{figure}[htbp]
	\centering
	\includegraphics[width=1.0\linewidth]{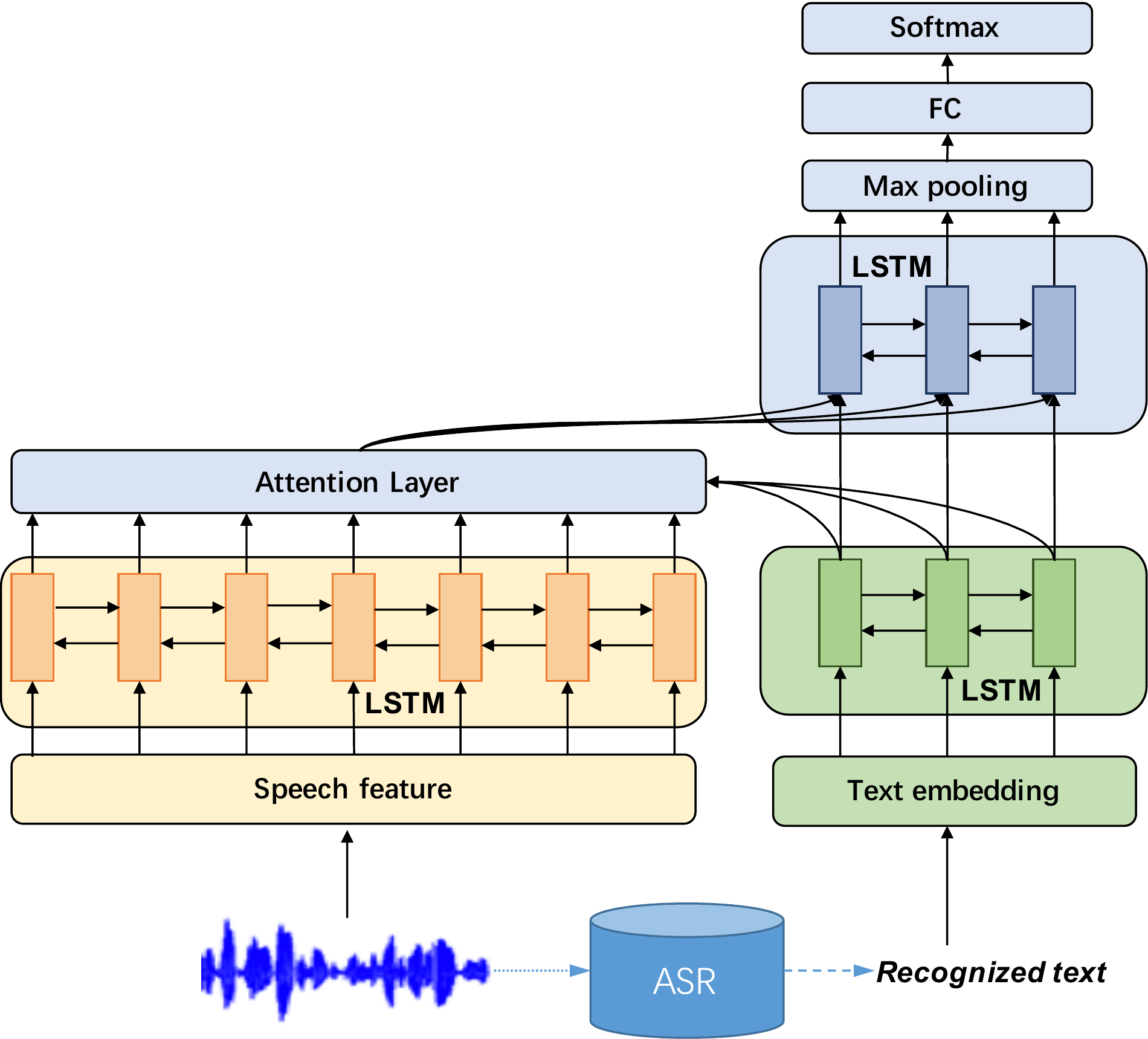}
	\caption{The architecture of the proposed model. The yellow part indicates the speech encoder and the green part indicates the text encoder. The blue part is the multimodal fusion network consisting of an attention network to fuse both modalities and an LSTM for sequence classification.}
	\label{fig:attention}
\end{figure}

The architecture of the model is shown in Figure \ref{fig:attention}. There are two paths to process a given speech signal. One path is to directly extract features from audio for speech encoding, and another path is to use an ASR system to produce text and covert to embedding for text encoding. Therefore the whole model consists of a speech encoder, a text encoder, and an multimodal fusion network including an attention mechanism and an LSTM for classification. We describe each component in detail in this section.

\subsection{Speech Encoder}
We first discuss the speech encoder in our multimodal emotion recognition model. To extract acoustic features, we first convert time-domain speech signals into frames with a 20 ms window and shifted every 10 ms. The low-level speech feature extracted from each frame can computed from the time-domain (e.g., zero-crossing rate), the spectral-domain (e.g., spectral spread), or the cepstral-domain (e.g., Mel-frequency cepstral coefficients, i.e., MFCC). We represent the sequence of features in an utterance as $[\bm x_1,...,\bm x_N]$, where $N$ is the number of frames in an utterance. 

For speech encoding, we choose a bidirectional LSTM (BiLSTM) to model the sequential structure of speech frames:
\begin{align}
&\overrightarrow{{\bm s}}_{i}=\overrightarrow{\text{LSTM}}({\bm x}_{i}), i\in\{1,..,N\}\\  
&\overleftarrow{{\bm s}}_{i}=\overleftarrow{\text{LSTM}}({\bm x}_{i}),i\in\{1,..,N\}\\
&{\bm s}_i=[\overrightarrow{{\bm s}}_{i}, \overleftarrow{{\bm s}}_{N-i+1}]
\end{align}
Here $\overrightarrow{\bm s_i}$ and $\overleftarrow{\bm s_i}$ are the hidden states of two unidirectional LSTMs, respectively. $\bm s_i$ is a concatenation of them, which will be used for alignment with text.

We mention that, although we do not focus on exploring speech encoders in this paper, we have experimented with various neural architectures similar to previous studies, such as CNN with LSTM \cite{satt2017efficient} and LSTM with attention \cite{mirsamadi2017automatic}. We observe comparable results for these architectures when combining with the proposed multimodal model.

\subsection{Text Encoder}
For emotion recognition of human speech, the speech can be translated to text with an ASR system. In our study, instead of training an ASR specific to the speech emotion recognition dataset, we use the public Google Cloud Speech API \footnote{https://cloud.google.com/speech-to-text/} to generate the text from speech, demonstrating the generalization of the proposed approach. Note that, our approach can tolerate some recognition errors and it is sufficient to train a model using these imperfect text. We will analyze the effects of ASR in Section 4.

Given a sequence of words, we first convert each word as an embedding vector $\bm e_j$, and the sequence is represented as $[{\bm e_{1},...,\bm e_{M}}]$, where $M$ is the number of words in the sentence. Then, we use a BiLSTM to model the text sequence. The hidden state $\bm h_j$ of the BiLSTM encodes the $j$th word in the sequence and will be used for further multimodal alignment. 
\begin{align}
    &\overrightarrow{{\bm h}}_{j}=\overrightarrow{\text{LSTM}}({\bm e}_{j}), j\in\{1,..,M\}\\  
    &\overleftarrow{{\bm h}}_{j}=\overleftarrow{\text{LSTM}}({\bm e}_{j}),j\in\{1,..,M\}\\
    &{\bm h}_j=[\overrightarrow{{\bm h}}_{j}, \overleftarrow{{\bm h}}_{M-j+1}]
\end{align}

\subsection{Attention Based Alignment}
An attention network was originally proposed in a sequence-to-sequence setting, where a decoder learns which parts in the encoder it should pay attention to and decode a word step by step \cite{bahdanau2014neural,chorowski2015attention}. In this study, instead of the decoding purpose, we utilize the attention mechanism to learn the alignment weights between speech frames and text words. This is similar to the self-attention approach in \cite{vaswani2017attention}, but the difference is that we learn the attention from two different sequences instead of the same sequence.

Specifically, an attention weight between the $i$th speech frame and the $j$th word is calculated by the hidden state $\bm h_j$ of the text LSTM and the hidden state $\bm s_i$ of the speech LSTM:
\begin{align}
&a_{j, i} = \tanh({\bm u}^\top{\bm s}_{i}+{\bm v}^\top {\bm h}_{j} + {b})\\
&\alpha_{j, i}=\frac{e^{a_{j,i}}}{\sum_{t=1}^N e^{a_{j, t}}} \\
&{\bm {\tilde s}}_{j}=\sum_{i}\alpha_{j, i}{\bm s}_i
\end{align}
where $\bm u$, $\bm v$ and $b$ are trainable parameters. $\alpha_{j,i}$ is the normalized attention weight over the speech sequence, indicating the soft alignment strength between the $j$th word and the $i$th speech frame. $\bm {\tilde s}_j$ is the weighted summation of hidden states from the speech LSTM, which is considered as an aligned speech feature vector corresponding to the $j$th word.

We then concatenate the aligned speech feature $\bm {\tilde s}_j$ and the hidden state of the text LSTM $\bm h_j$ to form a combined multimodal feature vector, which is fed into a multimodal BiLSTM for feature fusion:
\begin{align}
&\overrightarrow{{\bm c}}_{j}=\overrightarrow{\text{LSTM}}([\bm {\tilde s}_j, \bm h_j]), j\in\{1,..,M\}\\  
&\overleftarrow{{\bm c}}_{j}=\overleftarrow{\text{LSTM}}([\bm {\tilde s}_j, \bm h_j]),j\in\{1,..,M\}\\
&{\bm c}_j=[\overrightarrow{{\bm c}}_{j}, \overleftarrow{{\bm c}}_{M-j+1}]
\end{align}

For emotion classification on a sequence, we apply an max-pooling layer over all hidden states in the sequence to get a fixed-length vector and then use a fully-collected layer with rectified linear units (ReLUs) for non-linear transformation. The loss $\mathcal{L}$ for each example is computed using a softmax layer with cross entropy for $C$-class classification.
\begin{align}
    &\bm {\tilde c}=\text{max\_pooling}([\bm c_1,...,\bm c_M]) \\
    &\bm z=\bm \phi(\bm W^\top \bm {\tilde c}), \bm z\in\mathbb R^C \\
    &p_c=\frac{e^{z_c}}{\sum_{k=1}^C e^{z_k}} \\
    &\mathcal{L}=-\sum_{c=1}^C y_c \log p_c 
\end{align}
where $\bm W$ is a trainable weight matrix, $\bm\phi$ is a point-wise ReLU transformation, $z_c$ is the $c$th element in $\bm z$, and $y_c=1$ if the ground-truth label is $c$ else $0$.

\section{Evaluations}
We discuss the dataset, implementation details and experimental results in this section.

\subsection{Data}
We use the Interactive Emotional Dyadic Motion Capture database (IEMOCAP)\footnote{https://sail.usc.edu/iemocap/index.html}\cite{busso2008iemocap} for experiments. The dataset was recorded from ten actors, and divided into five sessions. Each dialog contains audio, transcriptions, video, and motion-capture recordings, and we only use audio in our study. There are both performances of improvisations and scripts of two different gender actors in a session. The recorded dialogues have been segmented into utterances and labelled as 10 categories (angry, happy, sad, neutral, frustrated, excited, fearful, surprised, disgusted, other). Each utterance was annotated by three different evaluators. In our experiments, we use four emotions (\textit{angry}, \textit{happy}, \textit{neutral} and \textit{sad}) for classification and use four sessions for model training and remaining for testing. This setting is consistent with prior studies. 

\subsection{Implementation}
For speech features, each utterance is sampled at 16 kHz with duration range from 0.5 to about 20 seconds. The time-domain signal is converted into 20 ms frames with 10 ms overlap. We use a Python library \cite{giannakopoulos2015pyaudioanalysis} to extract a 34-dimensional feature vector from each frame including MFCC, zero-crossing rate, spectral spread, spectral centroid, etc.

For text features, as we mentioned before, we first use Google Cloud speech service to generate the text from speech signals. Based on the text transcripts provided by the IEMOCAP dataset, the word error rate of Google speech service is 14.7\%. For word representation, we use a 300-dimensional GloVe embedding \cite{pennington2014glove} as the pretrained text embedding.

To implement the model, we use 100 hidden units in each unidirectional LSTM in the speech encoder, the text encoder, and the multimodal encoder, so the dimensionality of a hidden state in a BiLSTM is 200.  The attention network has 5 attention heads, each of which includes 40 weights. The fully-connected layer is a $200\times 4$ weight matrix corresponding to the number of hidden states and the number of classes. To train the model, we use Adam optimization with the learning rate of 0.001.

We adopt two widely used metrics for evaluation: weighted accuracy (WA) that is the overall classification accuracy and unweighted accuracy
(UA) that is the average recall over the emotion categories.

\subsection{Experiments}
For comparison, we first train models with each single modality separately. For speech modality, we use an LSTM to model the sequence of speech frames and use an attention mechanism to learn a weighted sum over the sequence. This structure is the same as in \cite{mirsamadi2017automatic} but with different speech features. 
Besides, we also report the results using CNN+LSTM in \cite{satt2017efficient} and TDNN+LSTM in \cite{sarma2018emotion} for comparison. 
Besides, we also report the results using CNN+LSTM in \cite{satt2017efficient} for comparison. 
For text modality, we employ an LSTM with attention structure which is the same as the text encoder in our approach.

We also compare our approach with other multimodal approaches. To combine speech and text, a straightforward way is to train an LSTM for each modality separately, and then use pooling or attention to  aggregate the hidden states to obtain a fixed-length vector for each sequence. The two vectors can be concatenated together for the sequence level classification. This ``Concat'' approach is similar to the method in \cite{yoon2018multimodal} but with different features, and we show the results in the paper for comparison.

\begin{table} 
\caption{Comparison results on the IEMOCAP dataset using speech-only, text-only, and multimodal models. All experiments in this table use recognized text. Bold fonts indicate the best performance.}
\centering
       \begin{tabular}{c c c}
       \hline \hline
    Methods & WA & UA  \\
    \hline 
    \textbf{{Speech-only}}  & & \\
      LSTM+Attn (our implementation) & 63.4 & 57.4 \\
      LSTM+Attn (Mirsamadi \textit{et al.}, 2017) & 63.5 & 58.8 \\
      CNN+LSTM (Satt \textit{et al.}, 2018) & 68 & 59.4 \\ 
       TDNN+LSTM (Sarma \textit{et al.}, 2018) & 70.1 & 60.7 \\ 
    \hline 
    \textbf{{Text-only (ASR text)}}  & & \\
      LSTM+Attn &60.3  & 54.8\\ 
    \hline 
    \textbf{{Multimodal (ASR text)}} & & \\
      Concat (our implementation) & 68.1 & 66.0 \\
      Concat (Yoon \textit{et al.}, 2018) & 69.1 & - \\
      Proposed & \textbf{70.4} & \textbf{69.5} \\
    \hline \hline
    \end{tabular}
    \label{tab:comp}
\end{table}

As shown in Table \ref{tab:comp}, ``LSTM+Attn'' in speech and ``LSTM+Attn'' in text are two unimodal models corresponding to our multimodal approach. By combining speech and recognized text, the multimodal approaches significantly boost both WA and UA. For comparison on multimodal methods, the proposed approach outperforms the direct concatenation approaches, showing the advantage of learned alignment between speech and text. We also report the results shown in other original papers and the proposed approach achieves the best results on both WA and UA.

Since the IEMOCAP provides text transcripts and word-level alignment, we conduct several experiments to analyze the influence. At first, We do not change the structure of the proposed model and only replace the recognized text by the transcripts. This is considered as an upper bound for the proposed approach as it uses the oracle text. Another experiment is to use the oracle text with provided alignment. With the word-level alignment, it is not necessary to use the attention mechanism. For each word in the text sequence, we simply average the hidden states in the speech LSTM in corresponding frames and concatenate it with the hidden state in the text LSTM. This is a version of hard alignment for the proposed approach. For comparison, we also use the oracle text to train a unimodal model and a concatenation model as in \cite{yoon2018multimodal}.

\begin{table}
\caption{Experiment results on the oracle text and alignment. The proposed approach with the recognized text is shown in the last row for reference. Bold fonts indicate the best performance.}
\centering
       \begin{tabular}{c c c c}
       \hline \hline
    Methods & WA & UA  \\
    \hline 
    \textbf{{Oracle text}} & & \\
    Text-only & 63.3 & 57.8 \\ 
    Concat (our implementation) & 71 & 67.7\\
    Concat (Yoon \textit{et al.}, 2018) & 71.8 & - \\ 
    Hard alignment & 71.5 & 68.6 \\
    Proposed & \textbf{72.5} & \textbf{70.9} \\
    \hline
    Proposed (ASR text) & 70.4 &69.5 \\
    \hline \hline
    \end{tabular}
    \label{tab:align}
\end{table}

Table \ref{tab:align} shows the results using provided transcripts and alignment. Comparing with the results in Table \ref{tab:comp}, the oracle text contributes around 3\% improvement for the text-only method and the direct concatenation method. Yoon \textit{et al.} \cite{yoon2018multimodal} also used the oracle text for experiment and achieved slightly better results than our implementation. The proposed approach with the oracle text achieves the best results in the dataset, showing that further improvement can be achieved by more accurate speech recognition. It is interesting to compare the proposed attention alignment with the hard alignment. Although the hard alignment approach utilizes the ground-truth alignment to aggregate the speech features, the performance is lower than the attention based method, suggesting that the attention network is optimized for emotion recognition rather than speech recognition.


\section{Conclusions}
In this paper, we aim to address emotion recognition from speech. With an ASR system, we can generate text from speech signals and build a multimodal model for emotion recognition. We propose an attention mechanism to learn the alignment between the original speech and the recognized text, which is then used to fuse features from two modalities. The fused features are fed into a sequence model for emotion classification. The experiment results show that the proposed approach is superior to other approaches in terms of emotion recognition results.
The experiments show that the proposed approach achieves state-of-the-art results on the dataset.

\bibliographystyle{IEEEtran}
\bibliography{./align_speech_text}

\begin{thebibliography}{10}
\providecommand{\url}[1]{#1}
\csname url@samestyle\endcsname
\providecommand{\newblock}{\relax}
\providecommand{\bibinfo}[2]{#2}
\providecommand{\BIBentrySTDinterwordspacing}{\spaceskip=0pt\relax}
\providecommand{\BIBentryALTinterwordstretchfactor}{4}
\providecommand{\BIBentryALTinterwordspacing}{\spaceskip=\fontdimen2\font plus
\BIBentryALTinterwordstretchfactor\fontdimen3\font minus
  \fontdimen4\font\relax}
\providecommand{\BIBforeignlanguage}[2]{{%
\expandafter\ifx\csname l@#1\endcsname\relax
\typeout{** WARNING: IEEEtran.bst: No hyphenation pattern has been}%
\typeout{** loaded for the language `#1'. Using the pattern for}%
\typeout{** the default language instead.}%
\else
\language=\csname l@#1\endcsname
\fi
#2}}
\providecommand{\BIBdecl}{\relax}
\BIBdecl

\bibitem{ngiam2011multimodal}
J.~Ngiam, A.~Khosla, M.~Kim, J.~Nam, H.~Lee, and A.~Y. Ng, ``Multimodal deep
  learning,'' in \emph{Proceedings of the 28th international conference on
  machine learning (ICML-11)}, 2011, pp. 689--696.

\bibitem{mikolov2013distributed}
T.~Mikolov, I.~Sutskever, K.~Chen, G.~S. Corrado, and J.~Dean, ``Distributed
  representations of words and phrases and their compositionality,'' in
  \emph{Advances in neural information processing systems}, 2013, pp.
  3111--3119.

\bibitem{russakovsky2015imagenet}
O.~Russakovsky, J.~Deng, H.~Su, J.~Krause, S.~Satheesh, S.~Ma, Z.~Huang,
  A.~Karpathy, A.~Khosla, M.~Bernstein \emph{et~al.}, ``Imagenet large scale
  visual recognition challenge,'' \emph{International journal of computer
  vision}, vol. 115, no.~3, pp. 211--252, 2015.

\bibitem{chorowski2015attention}
J.~K. Chorowski, D.~Bahdanau, D.~Serdyuk, K.~Cho, and Y.~Bengio,
  ``Attention-based models for speech recognition,'' in \emph{Advances in
  neural information processing systems}, 2015, pp. 577--585.

\bibitem{chan2016listen}
W.~Chan, N.~Jaitly, Q.~Le, and O.~Vinyals, ``Listen, attend and spell: A neural
  network for large vocabulary conversational speech recognition,'' in
  \emph{2016 IEEE International Conference on Acoustics, Speech and Signal
  Processing (ICASSP)}.\hskip 1em plus 0.5em minus 0.4em\relax IEEE, 2016, pp.
  4960--4964.

\bibitem{neiberg2006emotion}
D.~Neiberg, K.~Elenius, and K.~Laskowski, ``{Emotion recognition in spontaneous
  speech using GMMs},'' in \emph{Ninth International Conference on Spoken
  Language Processing}, 2006.

\bibitem{nogueiras2001speech}
A.~Nogueiras, A.~Moreno, A.~Bonafonte, and J.~B. Mari{\~n}o, ``Speech emotion
  recognition using hidden markov models,'' in \emph{Seventh European
  Conference on Speech Communication and Technology}, 2001.

\bibitem{mower2011framework}
E.~Mower, M.~J. Mataric, and S.~Narayanan, ``A framework for automatic human
  emotion classification using emotion profiles,'' \emph{IEEE Transactions on
  Audio, Speech, and Language Processing}, vol.~19, no.~5, pp. 1057--1070,
  2011.

\bibitem{stuhlsatz2011deep}
A.~Stuhlsatz, C.~Meyer, F.~Eyben, T.~Zielke, G.~Meier, and B.~Schuller, ``Deep
  neural networks for acoustic emotion recognition: raising the benchmarks,''
  in \emph{2011 IEEE international conference on acoustics, speech and signal
  processing (ICASSP)}.\hskip 1em plus 0.5em minus 0.4em\relax IEEE, 2011, pp.
  5688--5691.

\bibitem{kim2013emotion}
Y.~Kim and E.~M. Provost, ``Emotion classification via utterance-level
  dynamics: A pattern-based approach to characterizing affective expressions,''
  in \emph{2013 IEEE International Conference on Acoustics, Speech and Signal
  Processing}.\hskip 1em plus 0.5em minus 0.4em\relax IEEE, 2013, pp.
  3677--3681.

\bibitem{han2014speech}
K.~Han, D.~Yu, and I.~Tashev, ``Speech emotion recognition using deep neural
  network and extreme learning machine,'' in \emph{Fifteenth annual conference
  of the international speech communication association}, 2014.

\bibitem{lee2015high}
J.~Lee and I.~Tashev, ``High-level feature representation using recurrent
  neural network for speech emotion recognition,'' in \emph{Sixteenth Annual
  Conference of the International Speech Communication Association}, 2015.

\bibitem{mirsamadi2017automatic}
S.~Mirsamadi, E.~Barsoum, and C.~Zhang, ``Automatic speech emotion recognition
  using recurrent neural networks with local attention,'' in \emph{2017 IEEE
  International Conference on Acoustics, Speech and Signal Processing
  (ICASSP)}.\hskip 1em plus 0.5em minus 0.4em\relax IEEE, 2017, pp. 2227--2231.

\bibitem{sarma2018emotion}
M.~Sarma, P.~Ghahremani, D.~Povey, N.~K. Goel, K.~K. Sarma, and N.~Dehak,
  ``{Emotion identification from raw speech signals using DNNs},'' \emph{Proc.
  Interspeech 2018}, pp. 3097--3101, 2018.

\bibitem{li2018attention}
P.~Li, Y.~Song, I.~McLoughlin, W.~Guo, and L.~Dai, ``An attention pooling based
  representation learning method for speech emotion recognition,'' \emph{Proc.
  Interspeech 2018}, pp. 3087--3091, 2018.

\bibitem{trigeorgis2016adieu}
G.~Trigeorgis, F.~Ringeval, R.~Brueckner, E.~Marchi, M.~A. Nicolaou,
  B.~Schuller, and S.~Zafeiriou, ``Adieu features? end-to-end speech emotion
  recognition using a deep convolutional recurrent network,'' in \emph{2016
  IEEE international conference on acoustics, speech and signal processing
  (ICASSP)}.\hskip 1em plus 0.5em minus 0.4em\relax IEEE, 2016, pp. 5200--5204.

\bibitem{satt2017efficient}
A.~Satt, S.~Rozenberg, and R.~Hoory, ``Efficient emotion recognition from
  speech using deep learning on spectrograms.'' in \emph{INTERSPEECH}, 2017,
  pp. 1089--1093.

\bibitem{busso2004analysis}
C.~Busso, Z.~Deng, S.~Yildirim, M.~Bulut, C.~M. Lee, A.~Kazemzadeh, S.~Lee,
  U.~Neumann, and S.~Narayanan, ``Analysis of emotion recognition using facial
  expressions, speech and multimodal information,'' in \emph{Proceedings of the
  6th international conference on Multimodal interfaces}.\hskip 1em plus 0.5em
  minus 0.4em\relax ACM, 2004, pp. 205--211.

\bibitem{wollmer2010context}
M.~W{\"o}llmer, A.~Metallinou, F.~Eyben, B.~Schuller, and S.~Narayanan,
  ``Context-sensitive multimodal emotion recognition from speech and facial
  expression using bidirectional lstm modeling,'' in \emph{Proc. INTERSPEECH
  2010, Makuhari, Japan}, 2010, pp. 2362--2365.

\bibitem{poria2017review}
S.~Poria, E.~Cambria, R.~Bajpai, and A.~Hussain, ``A review of affective
  computing: From unimodal analysis to multimodal fusion,'' \emph{Information
  Fusion}, vol.~37, pp. 98--125, 2017.

\bibitem{yoon2018multimodal}
S.~Yoon, S.~Byun, and K.~Jung, ``Multimodal speech emotion recognition using
  audio and text,'' in \emph{IEEE SLT}, 2018.

\bibitem{zadeh2017tensor}
A.~Zadeh, M.~Chen, S.~Poria, E.~Cambria, and L.-P. Morency, ``Tensor fusion
  network for multimodal sentiment analysis,'' in \emph{Proceedings of the 2017
  Conference on Empirical Methods in Natural Language Processing}, 2017, pp.
  1103--1114.

\bibitem{poria2017context}
S.~Poria, E.~Cambria, D.~Hazarika, N.~Majumder, A.~Zadeh, and L.-P. Morency,
  ``Context-dependent sentiment analysis in user-generated videos,'' in
  \emph{Proceedings of the 55th Annual Meeting of the Association for
  Computational Linguistics (Volume 1: Long Papers)}, vol.~1, 2017, pp.
  873--883.

\bibitem{bahdanau2014neural}
D.~Bahdanau, K.~Cho, and Y.~Bengio, ``Neural machine translation by jointly
  learning to align and translate,'' in \emph{Proceedings of ICLR}, 2015.

\bibitem{vaswani2017attention}
A.~Vaswani, N.~Shazeer, N.~Parmar, J.~Uszkoreit, L.~Jones, A.~N. Gomez,
  {\L}.~Kaiser, and I.~Polosukhin, ``Attention is all you need,'' in
  \emph{Advances in Neural Information Processing Systems}, 2017, pp.
  5998--6008.

\bibitem{busso2008iemocap}
C.~Busso, M.~Bulut, C.-C. Lee, A.~Kazemzadeh, E.~Mower, S.~Kim, J.~N. Chang,
  S.~Lee, and S.~S. Narayanan, ``{IEMOCAP: Interactive emotional dyadic motion
  capture database},'' \emph{Language resources and evaluation}, vol.~42,
  no.~4, p. 335, 2008.

\bibitem{giannakopoulos2015pyaudioanalysis}
T.~Giannakopoulos, ``{pyAudioAnalysis: An open-source Python library for audio
  signal analysis},'' \emph{PloS one}, vol.~10, no.~12, 2015.

\bibitem{pennington2014glove}
J.~Pennington, R.~Socher, and C.~Manning, ``Glove: Global vectors for word
  representation,'' in \emph{Proceedings of the 2014 conference on empirical
  methods in natural language processing (EMNLP)}, 2014, pp. 1532--1543.

\end{thebibliography}
\end{document}